\documentclass[11pt]{article}

\usepackage[preprint]{acl}
\usepackage{pbalance}
\usepackage{times}
\usepackage{latexsym}
\usepackage{subfigure}
\usepackage[T1]{fontenc}
\usepackage{tcolorbox}
\usepackage{amsfonts}  
\usepackage[utf8]{inputenc}
\usepackage{booktabs}       
\usepackage{microtype}
\usepackage{algorithm}
\usepackage{algorithmic}
\usepackage{inconsolata}
\usepackage{makecell}
\usepackage{graphicx}
\usepackage{multirow}
\usepackage{enumitem}
\usepackage{color}
\usepackage{float}
\usepackage{amsmath}
\usepackage{amssymb}
\newtheorem{theorem}{Theorem}

\newcommand{\bA}{\mathbf{A}}
\newcommand{\bQ}{\mathbf{Q}}
\newcommand{\tcot}{\texttt{CoT}}
\newcommand{\tvf}{\texttt{VF}}
\newcommand{\bR}{\mathbf{R}}
\newcommand{\bS}{\mathbf{S}}

\title{Asking LLMs to Verify First is Almost Free Lunch}

\author{Shiguang Wu\\
	Department of Electonic Engineering \\
	Tsinghua University \\
	\texttt{wsg23@mails.tsinghua.edu.cn} \\\And
	Quanming Yao \\
	Department of Electonic Engineering \\
	Tsinghua University \\
	\texttt{qyaoaa@tsinghua.edu.cn} \\}

\begin{document}
	\maketitle
	\begin{abstract}
		To enhance the reasoning capabilities of Large Language Models (LLMs) without high costs of training, nor extensive test-time sampling, we introduce Verification-First (VF), a strategy that prompts models to verify a provided candidate answer, even a trivial or random one, before generating a solution. This approach triggers a "reverse reasoning" process complementary to standard forward Chain-of-Thought (CoT), which restricts the logical search space of the answer by pruning the LLM's output distribution. We further generalize VF prompting to Iter-VF, a sequential test-time scaling (TTS) method that iteratively cycles the verification-generation process using the model's previous answer. Extensive experiments across various benchmarks and various LLMs confirm that VF prompting with random answer consistently outperforms standard CoT with minimal computational overhead, and Iter-VF outperforms existing TTS strategies. VF is also effective on SOTA thinking models. For example, by using the simple VF prompting, we obtain a new SOTA $94.9\%$ accuracy on GPQA-Diamond with Gemini-3-Pro-Preview where VF reduces its errors by $\sim30\%$ relatively. Code is provided at \url{https://anonymous.4open.science/r/VF-Code-DDD2}.
	\end{abstract}
	
	\section{Introduction}
	To make LLMs adept at complex reasoning tasks,
	it is common to convert a complex problem into multi-step, modular and primary reasoning steps within their capacity. 
	A fundamental technique is to ask the LLM to "think step by step", forming CoT \cite{wei2022chain}.
	Though generating such a reasoning path to the final solution would be much more simpler than directly output the solution,
	their reliability is still undermined by their tendency to generate plausible but incorrect solutions, including hallucination and error propagation. This fallibility stems from their pre-trained nature to generate coherent natural language auto-regressively word by word, but the lack of logical reasoning, which can prioritize fluency over factual or logical rigor.

	To enhance LLM reasoning, existing methods 
	take CoT process as primary components, to develop strategies beyond,
	with significant costs at three dimensions: prior knowledge, test-time computation, and training. Strategies typically involve crafting problem-specific prompts \cite{wei2022chain,chia2023contrastive,alazraki2025no}, increasing inference budgets through expensive parallel sampling \cite{wang2022self} or sequential reflection \cite{madaan2023self,shinn2023reflexion}, fine-tuning models \cite{cobbe2021training,kumar2025llm}, or involving multiple above perspectives \cite{yao2023tree,lightman2023let,besta2024graph,snell2024scaling,muennighoff2025s1,setlur2025scaling}. 
	This suggests a prevailing understanding: \textit{better reasoning can only be attained at a significant cost.}

	\begin{figure}[t]
		
		\centering
		\includegraphics[width=0.49\textwidth]{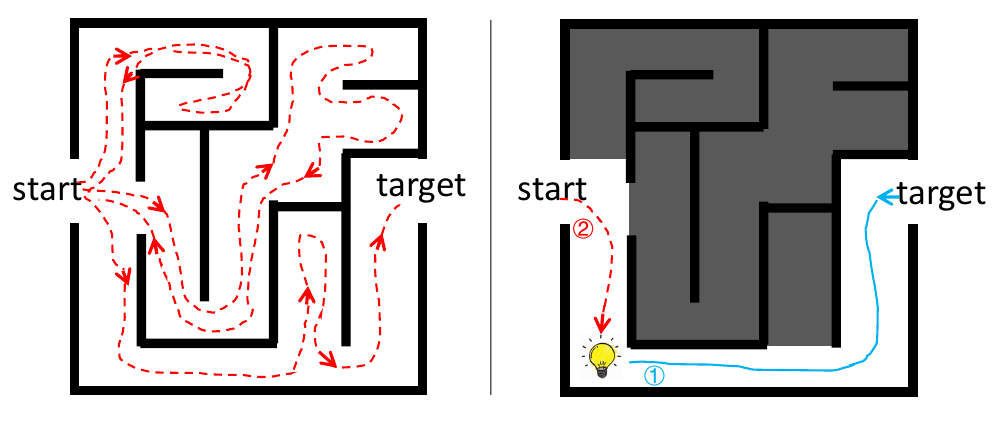}
		\vspace{-0px}
		\caption{The verification process (blue) restricts the logical search space of forward reasoning process (red).}
		\vspace{-10pt}
		\label{fig:1}
		\vspace{-0px}
	\end{figure}

		\begin{figure*}[t]
		\centering
		\vspace{-10pt}
		\includegraphics[width=0.95\textwidth]{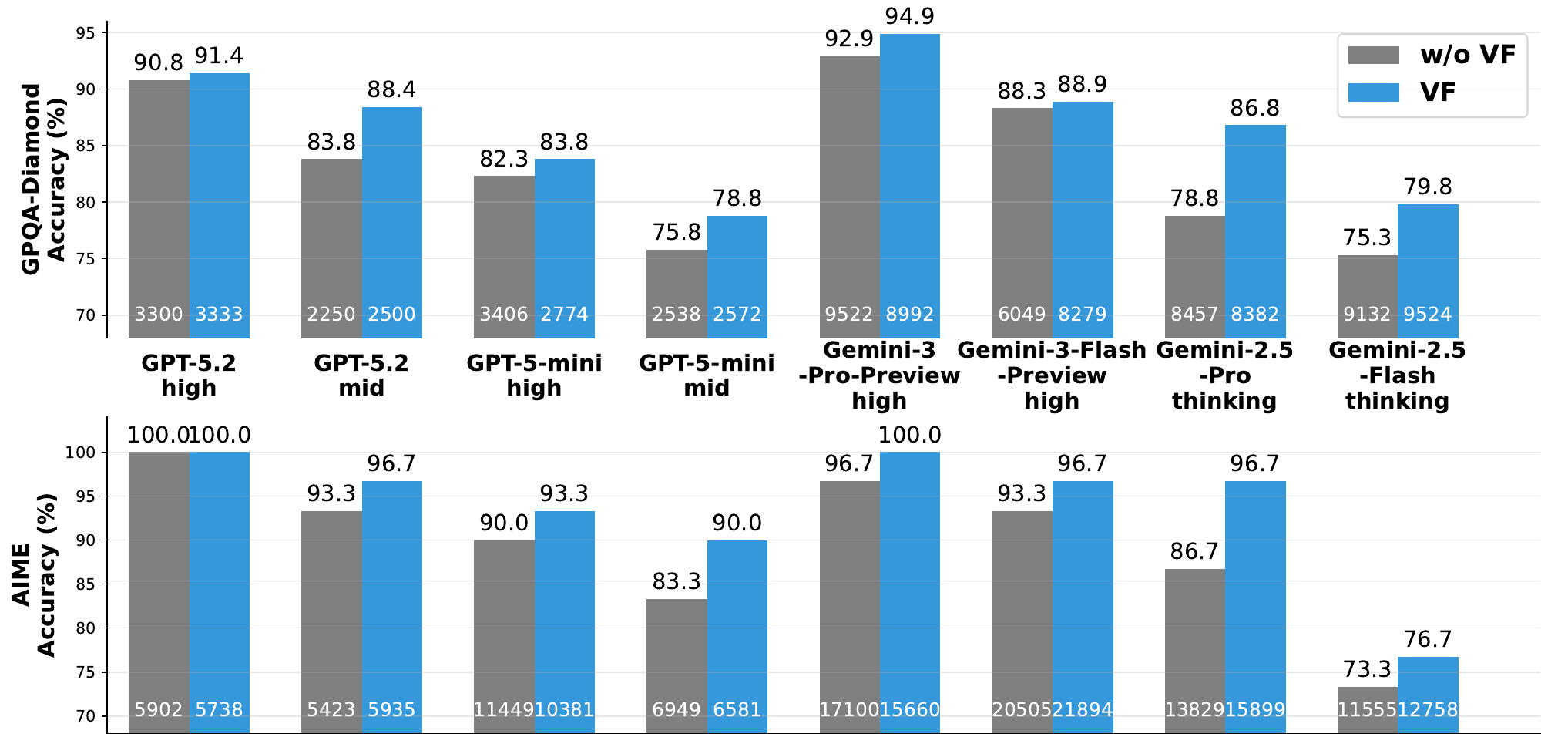}
		\vspace{-5pt}
		\caption{Performance comparison of thinking models between VF (prompting with random/trivial answer) and without (w/o) VF. The white number at bottom of each bar represent average token consumption.}
		\label{fig:perfomance0}
		\vspace{-10pt}
	\end{figure*}

	In this paper,
	we propose a method that is extremely cheap to improve LLM reasoning, by complementing the CoT process from the start.
	The core is Verification-First (VF) strategy, providing an answer 
	(regard its correctness or not)
	along with the problem and ask LLM to first verify/evaluate the provided answer then give correct answer, 
	in contrast to ordinary reasoning that starting from the problem only.
	The key insight is that verifying an answer restricts the logical search space of the final answer by pruning the LLM's output distribution autoregressively, with theoretical analysis provided.
	The VF strategy can be implemented by providing a random/trivial answer with minimal prior knowledge, or previously generated answer, and thus can be generalized as a TTS method.
	It turns out to improve reasoning with costing \textbf{minimal/zero prior knowledge, 
		no training, and minimal test-time computation} (comparing with existing TTS methods).
	For example shown in Figure~\ref{fig:perfomance0}, adding one simple sentence in the prompt can effectively improve the performance of SOTA thinking models, with almost no additional cost.
	Our contributions are:

	\begin{itemize}[leftmargin=*]
		
		\item We propose Verification-First strategy to
		improve LLM reasoning with low cost, by triggering reverse reasoning to restrict logical search space.
		
		\item We show VF strategy is easy to implement with existing LLMs,
		and is applicable in usages from zero-shot prompting to TTS.
		
		\item Extensive experiments show that the proposed algorithms outperforms standard CoT and existing TTS methods across various tasks and models.
	\end{itemize}

	
	\section{Related Works}
	
	\label{sec:rw1}
	
	To address the fallibility of LLMs generating coherent natural language, which can prioritize fluency over factual or logical rigor, 
	many methods tries to improve LLM reasoning ability beyond CoT.
	Existing methods typically impose additional costs from three distinct perspectives: prior knowledge, 
	test-time computation, and training.
	
	Some approaches depend heavily on task-specific customization. These methods require humans to provide extensive prior knowledge to craft prompts with more few-shot examples or delicate, task-specific instructions \cite{wei2022chain,chia2023contrastive,alazraki2025no}. This limits generalization as the prompt must be tailored to the specific problem.
	Another prominent line of work increases inference costs to make reasoning more deliberate. This is often achieved through parallel strategies, such as generating multiple candidates and voting \cite{wang2022self} or selecting the best via a reward model \cite{lightman2023let}. Alternatively, sequential strategies iteratively reflect on and refine previous steps \cite{madaan2023self,shinn2023reflexion}. More complex strategies combine these by decomposing steps into trees or graphs  \cite{yao2023tree,besta2024graph}. Recent studies on TTS suggest that significant performance gains in this paradigm generally require a proportional increase in token generation, meanwhile the evaluator or LLM itself should be correspondingly trained to be capable with the scaling strategy \cite{snell2024scaling,muennighoff2025s1,setlur2025scaling}.
	Finally, it has been widely accepted that model performance can be improved by fine-tuning on high-quality reasoning data \cite{cobbe2021training,kumar2025llm}. However, this incurs substantial training costs and is restrictive, as it requires access to open-source parameters and curated datasets, which is not feasible for all applications or closed-source models.
	
	\begin{figure*}[t]
		\centering
		\vspace{-0pt}
		\includegraphics[width=0.85\textwidth]{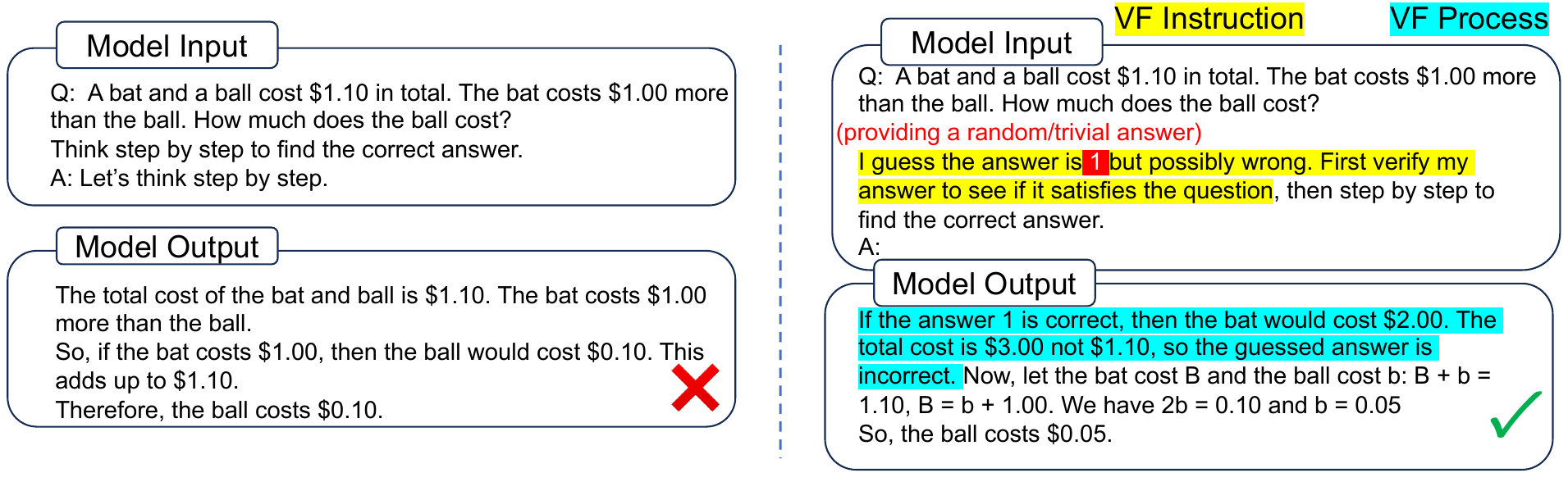}
		\caption{VF prompting with random/trivial answer (Right), comparing with standard CoT prompting (Left)}
		\label{fig:case0}	
		\vspace{-10pt}
	\end{figure*}

	\section{Proposed Method}

	
	The main idea of VF strategy is simple, 
	complementing CoT reasoning.
	Considering we have a problem $\bQ$ for LLM to answer, while the standard CoT instructs the LLM by $\tcot(\bQ):\simeq$
	"\textit{Think step by step to find the answer of $\bQ$}",
	VF instructs the LLM by $\tvf(\bQ,\bA'):\simeq$ which is defined as
	\begin{tcolorbox}
		$\tvf(\bQ,\bA'):\simeq$ ``\textit{A possible answer of $\bQ$ is $\bA'$. First verify if $\bA'$ is correct, then think step by step to find the answer}.''
	\end{tcolorbox}
	It is expected that the LLM would output to first verify the provided answer as a critic, like bringing back to the question to see if it satisfies conditions, and then reaches the final answer.
	We discuss two different ways to provide
	$\bA'$, corresponding to two scenarios to apply the method: zero-shot prompting (Section~\ref{sec:prompting}) and test-time scaling (Section~\ref{sec:tts}).
	
	\subsection{VF Prompting: Zero-Shot Prompting}\label{sec:prompting}
	With minimal cost, $\bA'$ can be random (or trivial).
	As we expect VF inherently takes advantage from the verification process, rather than the information gain in $\bA'$.
	For problems with simple answer space, where a random answer can be determined with minimal prior knowledge, while being nontrivial to verify, 
	user can provide such an answer as $\bA'$. 
	For example, For instance, a general heuristic is to $\bA'=1$ for math word problems and $\bA'=$"\textit{Option B}" for choice problems. The algorithm is described in Algorithm~\ref{alg:1}., and an illustrative case is provided in Figure~\ref{fig:case0}.
	

	We now establish a formal framework grounded in \textbf{constraint satisfaction} and \textbf{search-space restriction}. To ensure this formalization is self-consistent, we first explicitly state the assumptions governing the underlying capacities and behaviors of the LLM.
	
	\subsubsection{Theoretical Analysis}
	A problem $\bQ$ can be formalized as a set of constraints $C^*= \{c_1, c_2, \dots, c_k\}$. A valid solution must satisfy all constraints simultaneously. Thus, the true correct answer space $S^*$ is the intersection of the sets of answers satisfying each individual constraint:
	$S^* = \bigcap_{c_i\in C^*}^k S(c_i)$
	where $S(c_i)$ represents the space of all possible answers that satisfy constraint $c_i$. 
	
	Consider the following three core behavioral properties of LLMs, which are inherent to human-like intelligence:
	\textbf{Assumption 1 (Limited Capacity):} The LLM's attention is inherently bounded. It naturally prioritizes explicit, lexically salient, or frequent constraints while failing to attend to implicit or logically complex constraints.
	\textbf{Assumption 2 (Verification Asymmetry):} Verifying an candidate answer against the problem is easier for an LLM than generating the correct answer from scratch (the majority of people agree P$\neq$NP and LLMs are aligned with majority people). If the candidate is incorrect, this verification acts as a ``collision test'' that explicitly surfaces constraints violated by the candidate.
	\textbf{Assumption 3 (Contextual Adherence):} Once a constraint is explicitly verbalized in the model's immediate context window, the model's subsequent autoregressive decoding probability mass is strictly conditioned upon it.
	%
	Now given problem $\bQ$, denote all answers possibly generated by CoT as $\bA_{\text{CoT}}\in S_{\text{CoT}}$, and all possible answers possibly generated by VF as $\bA_{\text{VF}}\in S_{\text{VF}}$. We have the following theorem:
	\begin{theorem}[Search-Space Restriction via VF]
		$S^*\subseteq S_{\text{VF}} \subseteq S_{\text{CoT}}$, and $P(\mathbf{A}_{\text{VF}} \in S^*) \geq P(\mathbf{A}_{\text{CoT}} \in S^*).$
	\end{theorem}
	This means VF improves reasoning by restricting the search space of possible answers, proved and illustrated as following.

	\subsubsection{Search Space Restriction via Verification}
	
	In standard CoT, the LLM generates a reasoning path autoregressively. Due to Assumption 1, its inherently bounded attention during open-ended generation, the model often fails to instantiate all necessary constraints, actively operating on only a subset of constraints $C_{\text{CoT}} \subseteq C^*$. Consequently, standard CoT operates within an effectively larger, under-constrained search space:
	$
	S_{\text{CoT}} = \bigcap_{c_i \in C_{\text{CoT}}} S(c_i)
	$
	Because $C_{\text{CoT}} \subseteq C^*$, it follows that $S_{\text{CoT}} \supseteq S^*$. 
	
	The core mechanism of VF is to introduce a candidate answer $\bA'$ and instruct the model to verify it {before} the CoT process. 
	When instructed to verify $A'$, the model acts as a critic. 
	Consider Assumption 2, there are two circumstances: 
	{If $\bA'$ happens to be correct}, then LLM acknowledge it is correct with higher chance than generating the correct answer from scratch.
	{If $\bA'$ is incorrect}, $A' \notin S^*$, which is the usual case with random/trivial $\bA'$. It will violate at least one constraint $c_j\in C^*$. Following Assumption 2, verifying $\bA'$ acts as collision test, explicitly triggering the instantiation of at least one previously ignored constraint $c_j \in C^*$.
	Meanwhile the generated verification trace explicitly articulates $c_j$ and adds it to the context window (e.g., ``\textit{If the answer is $A'$, then condition $c_v$ would not hold\dots}'') before CoT process. Following {Assumption 3}, for the subsequent CoT generation phase, the model's active constraint set is expanded:
	$
	C_{\text{VF}} = C_{\text{CoT}} \cup \{c_j\}
	$
	Consequently, the effective search space for the final answer is restricted:
	$
	S_{\text{VF}} = S_{\text{CoT}} \cap S(c_j)
	$
	So we have $S^*\subseteq S_{\text{VF}} \subseteq S_{\text{CoT}}$, i.e., the space of plausible-but-incorrect reasoning paths is pruned. Thus for a answer $\bA_{\text{VF}}\in S_{\text{VF}}$, and a answer $\bA_{\text{CoT}}\in S_{\text{CoT}}$, we have
	$P(\bA_{\text{VF}}\in S^*)\geq P(\bA_{\text{CoT}}\in S^*)$, i.e., the answer generated by VF prompting has greater chance to be correct than CoT.
	
	
	\subsubsection{Example Illustration}
	
	To make this formalization concrete, consider the case shown in Figure~\ref{fig:case0}.
	\paragraph{The Constraints ($C$):}
	$c_1$: Total cost is \$1.10 ($\text{Bat} + \text{ball} = 1.10$).
	$c_2$: Bat is \$1.00 more than the ball ($\text{Bat} = \text{ball} + 1.00$).
	
	\paragraph{CoT:}
	The model focuses on the numbers (\$1.10 and \$1.00) and the operation of subtraction ($c_1$), while failing to rigorously instantiate the relative difference ($c_2$). It searches in an under-constrained space ($C_{\text{CoT}} = \{c_1\}$) and easily outputs $\text{ball} = 0.10$.
	
	\paragraph{VF:}
	We supply a trivial answer $A' = 1$. To check this, the model substitutes $\text{ball} = 1$ into the constraints. It computes $\text{Bat} = 2.00$ ($c_2$) and checks $\text{Total} = 3.00 \neq 1.10$ ($c_1$). This ``collision'' forces the model to explicitly write out the simultaneous relationship of both $c_1$ and $c_2$ in the context.
	The search space is now restricted by both constraints explicitly. The naive heuristic path ($0.10$) is pruned from the valid continuation space, virtually guaranteeing the model finds the correct algebraic intersection ($0.05$).

	\subsection{Iter-VF: Test-Time Scaling}\label{sec:tts}
	While zero-shot VF prompting is effective, it faces two limitations: defining a meaningful initial answer for complex, open-ended tasks is challenging, and it lacks a mechanism for controllable test-time scaling (TTS). To address these issues, we propose Iter-VF.
	
	\paragraph{Generating the Initial Answer.} For complex, open-ended tasks like coding, a trivial guess (e.g., $\bA'=$"\textit{print('Hello World')"}) provides little verification signal. To supply a non-trivial candidate without relying on prior knowledge, we simply ask the LLM to generate it. We first prompt the LLM using standard CoT to obtain an initial answer $\bA_1$
	, and then apply VF prompting, i.e., $\tvf(\bQ,\bA_1)$ to yield a final answer $\bA_2$.
	This two-call approach can be applied to any complex problem (Section~\ref{sec:open}).

	\paragraph{Iterative Test-Time Scaling.} To achieve controllable TTS, we generalize this process into a sequential iteration. In each step $i$, a VF process takes an old answer $\bA_{i-1}$ as input and generates an improved answer $\bA_i$. We repeat this cycle up to a defined computation budget $B$, as outlined in Algorithm~\ref{alg:3}. The initial candidate $\bA_0$ can be either a user-provided trivial answer or one generated via standard CoT.
	
	\begin{algorithm}[t]            
		\small  
		\caption{VF Prompting (zero-shot prompting)}
		\label{alg:1}
		\begin{algorithmic}
			\REQUIRE LLM $M$, problem $\bQ$, a random/trivial answer $\bA'$.
			\STATE Get final answer $\bA$ from $M(\tvf(\bQ,\bA'))$;
			\RETURN  answer $\bA$.
		\end{algorithmic}
	\end{algorithm}
	
	\begin{algorithm}[t]            
		\small  
		\caption{Iter-VF (TTS)}
		\label{alg:3}
		\begin{algorithmic}
			\REQUIRE LLM $M$, problem $\bQ$, computation budget $B$, (optional: initial answer $\bA_0$).
			\FOR{$i=1,\cdots,,B$}
			\STATE Get final answer $\bA_{i}$ from $M(\tvf(\bQ,\bA_{i-1}))$;
			\STATE Early stop if $\bA_{i}==\bA_{i-1}$;
			\ENDFOR
			\RETURN  answer $\bA_B$.
		\end{algorithmic}
	\end{algorithm}

	As a sequential TTS strategy, Iter-VF seems similar with existing strategies Self-Correction / Refine~/ Reflexion in the iterations. In fact they are very different, as discussed in Appendix~\ref{sec:framework_differentiation}. Iter-VF distinguishes itself by (i) maintaining a Markovian process across iterations, that avoids context overflow and error accumulation; and (ii) instructing to do reasoning from scratch and benefit from the verification process in every iteration.

	\begin{figure}[t]
		\centering
		\vspace{-10px}
		\includegraphics[width=0.48\textwidth]{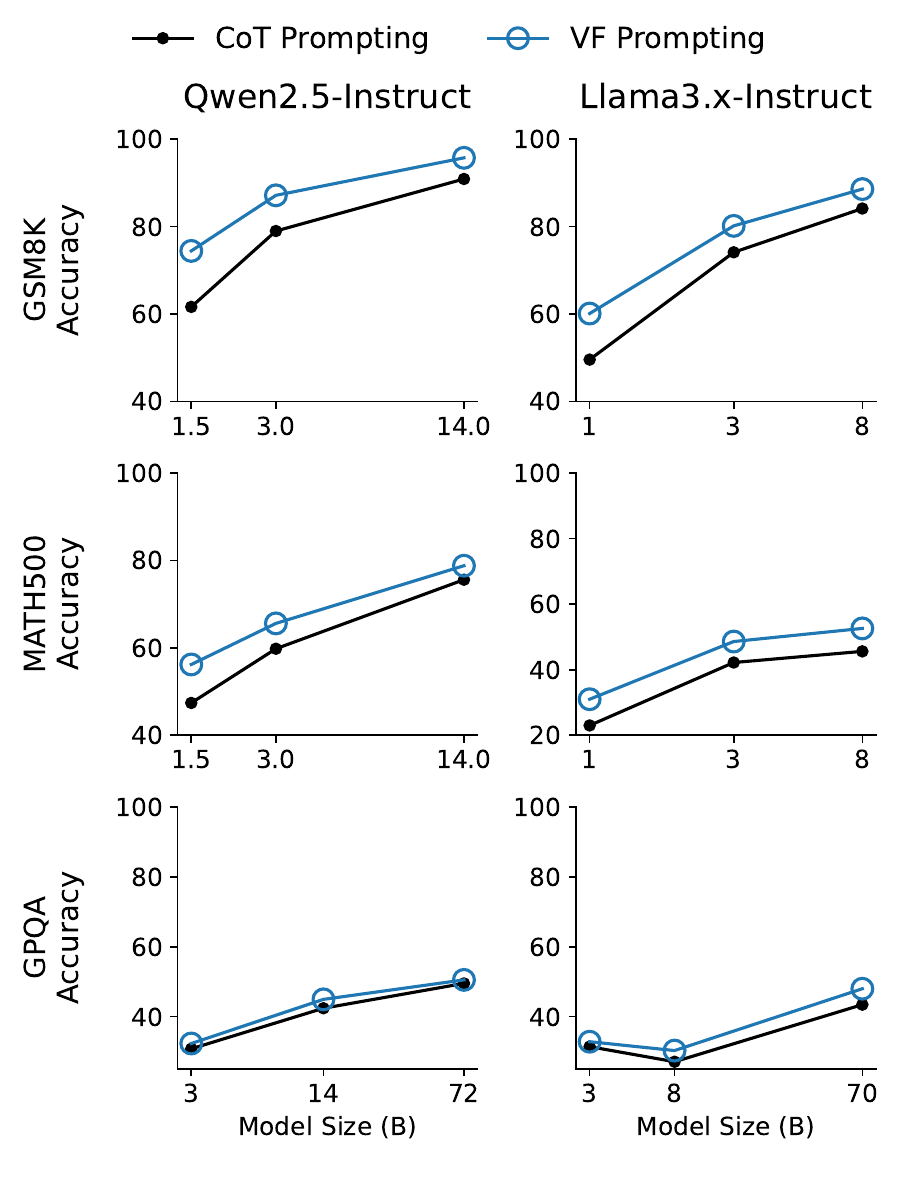}
		
		\vspace{-1px}
		\caption{VF prompting consistently outperforms standard CoT prompting. }
		\vspace{-10px}
		\label{fig:comp}
	\end{figure}

	\section{Experiments}
	The experiment evaluation includes the following main results: 
	(i) VF prompting with random/trivial answer shows consistent improvement over standard CoT prompting.
	(ii) Iter-VF outperforms existing TTS strategies under limited test-time computation budget on models without additional training.
	(iii) On open-end tasks in real-world scenarios, VF prompting with previous generated answer performs the best comparing with existing strategies using similar computation budget.
	(iv) With thinking models and thought-hidden LLM service where other strategies might be inapplicable or duplicate, VF strategy is still effective.
	Code will be provided to public.


	\subsection{VF Prompting with Random/Trivial Answer}\label{sec:exp-random}
	Adapting VF prompting with random/trivial answer  requires the random/trivial answer can be easily defined. 
	Most reasoning benchmark satisfies such requirement. We use reasoning benchmarks of math problems GSM8K \cite{cobbe2021training} and MATH500 \cite{hendrycks2021measuring,lightman2023let}, and a graduate-level science Q\&A benchmark GPQA-Diamond \cite{rein2024gpqa}. 
	We compare with standard CoT prompting (0-shot), as
	they are generally applicable costing comparable minimal prior knowledge. We provide trivial answer "1" for all problems in GSM8K and MATH500, and random choice of shuffled options for all problems in GPQA-Diamond.
	
	\begin{table}[t]
		\centering
		\small
		\begin{tabular}{c|c|c|c}
			\toprule
			Dataset&  GSM8K&MATH500& GPQA \\\midrule
			CoT&365.6&808.3&739.3\\
			VF&533.6&1109.6&901.8\\
			\bottomrule
		\end{tabular}
		\vspace{-0px}
		\caption{Numbers of output tokens.}
		\vspace{-10px}
		\label{tab:token}
	\end{table}
	We evaluate LLMs from the Qwen2.5 \cite{qwen2} and Llama3 \cite{grattafiori2024llama} families, covering size from 1B to 72B: 
	Qwen2.5-1.5B / 3B / 14B / 72B-Instruct, Llama3.2-1B / 3B-Instruct, LLama3.1-8B-Instruct, Llama3.3-70B-Instruct.

	\paragraph{Performance and Cost Comparison.}
	Figure~\ref{fig:comp} shows the results, and Appendix~\ref{app:num} provides details. We can find that VF consistently outperforms CoT, 
	with performance advantage being stable across different model sizes. 
	We notice the advantage on GSM8K and MATH500 is much more significant than GPQA-Diamond. 
	This could be interpreted by the intuition and fact \cite{zhao2025test} that facilitating reasoning improve LLMs' performance on knowledge-intensive problems much more hardly than computation/logic-intensive problems.

	For test-time computation, Table~\ref{tab:token} summarizes the numbers of output tokens to reach the final answer, 
	on the three benchmarks respectively averaged over the tested models. 
	Comparing with standard CoT prompting, VF prompting outputs about 20\%$\sim$50\% more tokens to make verification first. 
	Such additional test-time computation cost could be considered minimal, 
	comparing with other methods to achieve similar improvement,
	which would be shown in Section~\ref{sec:exp-tts}.

	\begin{figure*}[t]
		\centering
		\vspace{-0pt}
		\includegraphics[width=0.9\textwidth]{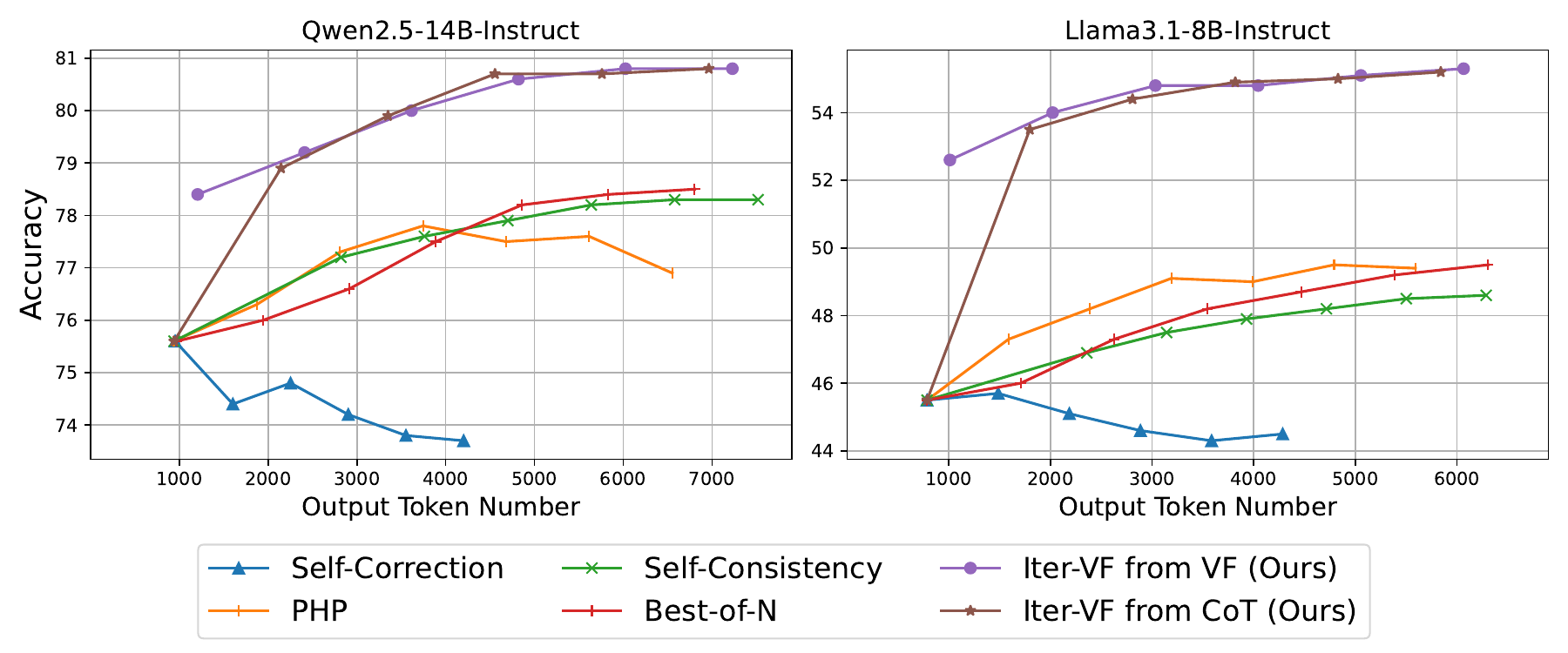}
		\vspace{-10pt}
		\caption{Comparison of different TTS methods on MATH500. }
		\vspace{-10pt}
		\label{fig:tts}	
	\end{figure*}

	Appendix~\ref{app:sensitive} provides Empirical results to supportLLMs verifying an answer is easier than generating the correct answer and providing different answers with minimal prior knowledge would not effect the final performance of VF prompting.  Appendix~\ref{sec:error-analysis} provides error case analysis where CoT gets correct but VF gets wrong.

	\subsection{Iter-VF for Test-Time Scaling}\label{sec:exp-tts}
	
	\paragraph{Baselines.}
	Since Iter-VF does not require additional training, external supervision, nor task-specific prior knowledge to decompose the problem, 
	we compare Iter-VF with TTS strategies including sequential ones: \textit{Self-Correction} \cite{madaan2023self,shinn2023reflexion,muennighoff2025s1}, \textit{PHP} \cite{zheng2023progressive}; and parallel ones: \textit{Self-Consistency} \cite{wang2022self}, \textit{Best-of-N}  \cite{lightman2023let,yao2023tree}.
	The detail implementation and discussion are provided in Appendix~\ref{app:tts}.
	%
	%
	%
	We implemented two variants of Iter-VF:\textit{ Iter-VF from VF} which starts VF prompting with the trivial answer "1" in the initial iteration; and\textit{ Iter-VF from CoT} which uses standard CoT prompting in the initial iteration and start verification from the second iteration.

	\paragraph{Performance Comparison and Analysis.}
	Figure~\ref{fig:tts} shows the results.
	It can be observed that Iter-VF significantly outperforms other baselines with limited test-time computation budget. 
	
	Comparing between Iter-VF from VF and from CoT, the initial difference but fast catching up indicates what matters is the verification process rather than the initially provided answer.
	This observation can also explain the advantage of Iter-VF over PHP, which only provides previous answers as "near" hints, that LLM could not necessarily make effective use of them.
	Another important difference between Iter-VF and the other two sequential TTS strategies (PHP, Self-Correction), is that they consider accumulated information through the iterations, while Iter-VF is Markovian as it only considers the answer of last one output.
	The accumulation of information can have negative effect for LLMs, as they have limited capacity for long context, and hallucinate a lot. Especially, Self-Correction considers not only previous answers, but also entire thinking process. It's performance become catastrophic with the accumulation of history. This observation is in line with previous study \cite{huang2023large}. Though such method can be effective after specifically fine-tuning LLM with curated data \cite{snell2024scaling,muennighoff2025s1,setlur2025scaling}, it would be far from our general setting.

	As for parallel strategies (Self-Consistency, Best-of-N), they are conservative that do not utilize history information before determining final answer. Their performances improve stably with the growth of computation, though not significant. 
	Iter-VF outperforms them, and can be integrated with them as the following discussion.
	
	\paragraph{Discussion.}
	Note that the current implementation Iter-VF is consistent with Algorithm~\ref{alg:3}, generating one sequence, and evaluating final $\bA_B$ only (keeping consistent with all sequential strategies). 
	However, considering its unique fully Markov property, the simple early-stopping rule could be applied to Iter-VF: as soon as $\bA_i$ is identical with $\bA_{i-1}$, Iter-VF stops and returns the current answer. However, we disable this adaptive rule and use a fixed budget $B$ for all methods, because adaptive stopping changes the actual compute consumed by each method and introduces an additional design dimension beyond the update rule itself.
	With more computation budget, we suggest to simply modify the implementation of Iter-VF by combining with some parallel strategies: one can first generate the Iter-VF sequence and then determining the final answer by majority-voting among all answers occurred in the iterations ($\bA_1,\cdots,\bA_B$) to make full usage of them; or parallelize multiple Iter-VF paths independently.
	
	\subsection{Applying in Real-Wold Scenarios}
	In this section, we show the effect of proposed method in real-world scenarios beyond choice problem or math Q\&A. 
	We discuss two perspectives which users are likely to face:
	(i) facing open-ended applications, where it is difficult to define a random/trivial answer without prior knowledge; (ii) using thinking models and even thought-hidden LLM service, where the exact input prompt and output about the LLM thinking process are not accessible.

	\subsubsection{On Open-Ended Applications}\label{sec:open}
	\begin{table*}[t]
		\centering
		\small
		\begin{tabular}{c|c|c|c|c|c|c|c|c}
			\toprule
			Model&  \multicolumn{4}{c}{{Qwen2.5-14B-Instruct}}& \multicolumn{4}{c}{{Llama3.1-8B-Instruct}} \\\midrule
			Task& \multicolumn{2}{c}{{Coding}}& \multicolumn{2}{c}{{API-Bank}}& \multicolumn{2}{c}{{Coding}}& \multicolumn{2}{c}{{API-Bank}}\\\midrule
			Dataset&HEval & MBPP&Level-1&Level-2 &HEval & MBPP&Level-1&Level-2\\\midrule
			
			CoT (pass@1)&91.5&70.0&73.2&49.8&81.1&56.8&55.1&38.6\\\midrule
			
			Self-Correction ("pass@1")&95.1& 71.8 &85.0&64.4&85.9&56.2&63.9&55.6\\
			Iter-VF ("pass@1")&{96.9}&{74.8}&{85.4}&72.6&90.2&61.5&64.0&57.1\\\midrule
			
			CoT$\times$2 (pass@2)&94.5&75.9&76.7&55.2&84.8&62.6&56.6&41.2\\
			Self-Correction ("pass@2")&97.6&77.4&\textbf{86.8}&68.9&89.0&62.2&\textbf{65.5}&\textbf{59.5}\\
			Iter-VF ("pass@2")&\textbf{99.4}&\textbf{80.6}&\textbf{87.7}&\textbf{77.9}&\textbf{93.3}&\textbf{69.3}&\textbf{65.2}&\textbf{60.3}\\
			\bottomrule
		\end{tabular}
		\vspace{-0px}
		\caption{Performance comparison of different TTS methods (budget: 2 calls) on coding and API tasks. Rows of comparable results are not split by horizontal line.
			``HEval'' is short for ``HumanEval''.}
		\label{tab:real}
		\vspace{-5px}
	\end{table*}

	We consider the following coding and API tasks where VF with random/trivial answer is no longer applicable. 
	We evaluate VF prompting with previously generated answer (calling LLM twice, as the procedure in Figure~\ref{fig:case2}).

	\paragraph{Benchmarks.}
	For coding tasks, we evaluate on HumanEval \cite{chen2021evaluating} and MBPP \cite{austin2021program}. In such benchmarks, a coding problem is like "check if in given list of numbers, are any two numbers closer to each other than given threshold" or "write a function to find the volume of a sphere". 
	For API tasks, we evaluate on API-Bank \cite{li2023api} Level-1 and Level-2, which simulates a agentic scenario where LLM need to understand user's intention from dialogue, and refer to documentary containing API description of tool functions, to output a API call with correct API function and arguments. 
	
	\paragraph{Baselines}
	To make fair comparison with similar cost, we consider baselines calling the LLM twice:
	\begin{itemize}[leftmargin=*]
		\item CoT$\times$2 (pass@2): Call the LLM twice independently. The final answer would be considered correct if at least one answer passes all test cases.
		
		\item Self-Correction: In the second call, the thinking process and the answer output by the first call is provided along with the problem, and the LLM is asked to reflect and refine the answer.
	\end{itemize}
	No other parallel methods would be considered as such "pass@$k$" implementation is necessarily the best parallel methods can do with budget $k$. No other sequential methods would be considered, as under this setting without external feed-back, Reflexion \cite{shinn2023reflexion} is reduced to such implementation; PHP has only been designed for math problems.

	\paragraph{Performance.}
	Table~\ref{tab:real} shows the results. For sequential methods Self-Correction and Iter-VF, "pass@1" evaluates the success rate using the second output, and "pass@2" evaluates if at least one of the first and the second output is success, to be distinguished from evaluating two independent trials. 
	
	Comparing the final performance under the "pass@2" metric, Iter-VF performs the best. Note that on API-Bank tasks, especially Level-1, Self-Correction has very similar performance with Iter-VF. This should be considered with the fact that on such problems, the reasoning processes only take a small proportion of the context, much shorter than the problem input and output answer. In such cases, the factual difference between them is very little.
	Comparing the first three rows, the advantage of Self-Correction and Iter-VF (calling the LLM twice) over CoT (calling the LLM once) indicates additional test-time computation does benefit in most cases. Comparing the last three rows with the first three rows, twice the evolution chances makes considerable improvement, while Iter-VF (pass@1) still outperforms CoT$\times$2 (pass@2) in most cases.
	
	\subsubsection{With Thinking Models and Thought-Hidden LLM Service}
	
	Recent advanced LLMs are increasingly equipped with built-in "thinking modes," utilizing inherent reasoning and TTS strategies. Moreover, cutting-edge commercial models (e.g., GPT-5, Gemini 3) present a unique challenge: they are not only closed-source but also conceal their internal reasoning traces. Users face a {thought-hidden} problem where they can only provide input instructions and receive post-processed final answers. The exact prompts guiding the model and the intermediate reasoning tokens are completely obscured.
	
	
	\paragraph{Failure of Existing Prompting and TTS Methods.} This hidden thinking process is often extensive. For example, our evaluations on GPT-5 Nano/Mini reveal that internal reasoning consumes roughly 10$\times$
	more tokens than the visible output on benchmarks like MATH and GPQA. Because this process is opaque and the model's internal strategies are unknown, traditional prompting and TTS methods become either redundant or entirely invalid. For instance, appending a standard "think step by step" prompt is unlikely to yield further improvements if the model already does so natively. 
	We do not know if GPT-5 is already using trace-dependent methods like Self-Correction. Even if it has not, users cannot ask the LLM to reflect on or refine its reasoning process because they are denied access to the reasoning trace itself.
	
	\paragraph{Feasibility and Advantages of VF.}
	In contrast, the proposed VF strategy remains highly feasible and advantageous in this paradigm. Because VF only requires modifying the initial input instruction—asking the model to evaluate a provided candidate answer before proceeding—it bypasses the need to access, monitor, or explicitly edit the model's internal trace. The "first-verify-then-generate" formulation gives a clear, distinct directive that triggers a reverse-reasoning process. This is fundamentally distinguishable from the standard forward-reasoning strategies these models likely employ internally. Consequently, VF successfully steers and restricts the hidden reasoning space from the outset. As shown in Figure~\ref{fig:perfomance0}, VF effectively enhances the performance of powerful, thought-hidden models, bringing tangible accuracy improvement, and surprisingly, with hardly no additional cost.
	This could be interpreted as that
	by explicitly restricting the logical search space from the outset, VF prevents thinking models from wasting their naturally extensive internal reasoning tokens on dead-end trajectories, making the overall generation process more efficient and sometimes reducing total token consumption.

	\section{Conclusion}
	In this paper, we introduced Verification-First (VF), a simple but effective strategy to enhance LLM reasoning capabilities. By prompting models to verify a candidate answer, even a random or trivial one, before generating a solution, we effectively trigger a "reverse reasoning" process which restricts the logical search space of the answer by pruning the LLM's output distribution. We further extended this concept to Iter-VF, a sequential test-time scaling method that iteratively verifies and improves answers.
	Our experiments demonstrate that VF improves reasoning ability with minimal computational overhead, without requiring any model training nor task-specific prior knowledge. These findings suggest that the cognitive gap between verification and generation offers an "almost free lunch" for improving LLM reliability, establishing VF as a general-purpose, plug-and-play enhancement for modern reasoning models.

	\section{Limitations}
	Despite the effectiveness of VF and Iter-VF, our work has several limitations. 
	For open-ended or creative tasks where defining a "trivial" or "random" initial answer is semantically difficult, the zero-shot VF approach requires a preceding CoT step to generate a seed answer, increasing the initial cost. 
	While Iter-VF dominates in low-to-medium compute budgets, its sequential Markovian nature could means it may reach a performance ceiling earlier than the other non-Markovian TTS strategies, when the computation budget is virtually unlimited. Future work may explore optimal combinations of VF with parallel search algorithms to address this upper bound.
		\bibliography{vf}
		
		\clearpage
		\appendix

		\begin{figure*}[t]
			\centering
			\vspace{-10pt}
			\subfigure[VF prompting with previously generated answer.\label{fig:case2}]{
				\includegraphics[width=0.85\textwidth]{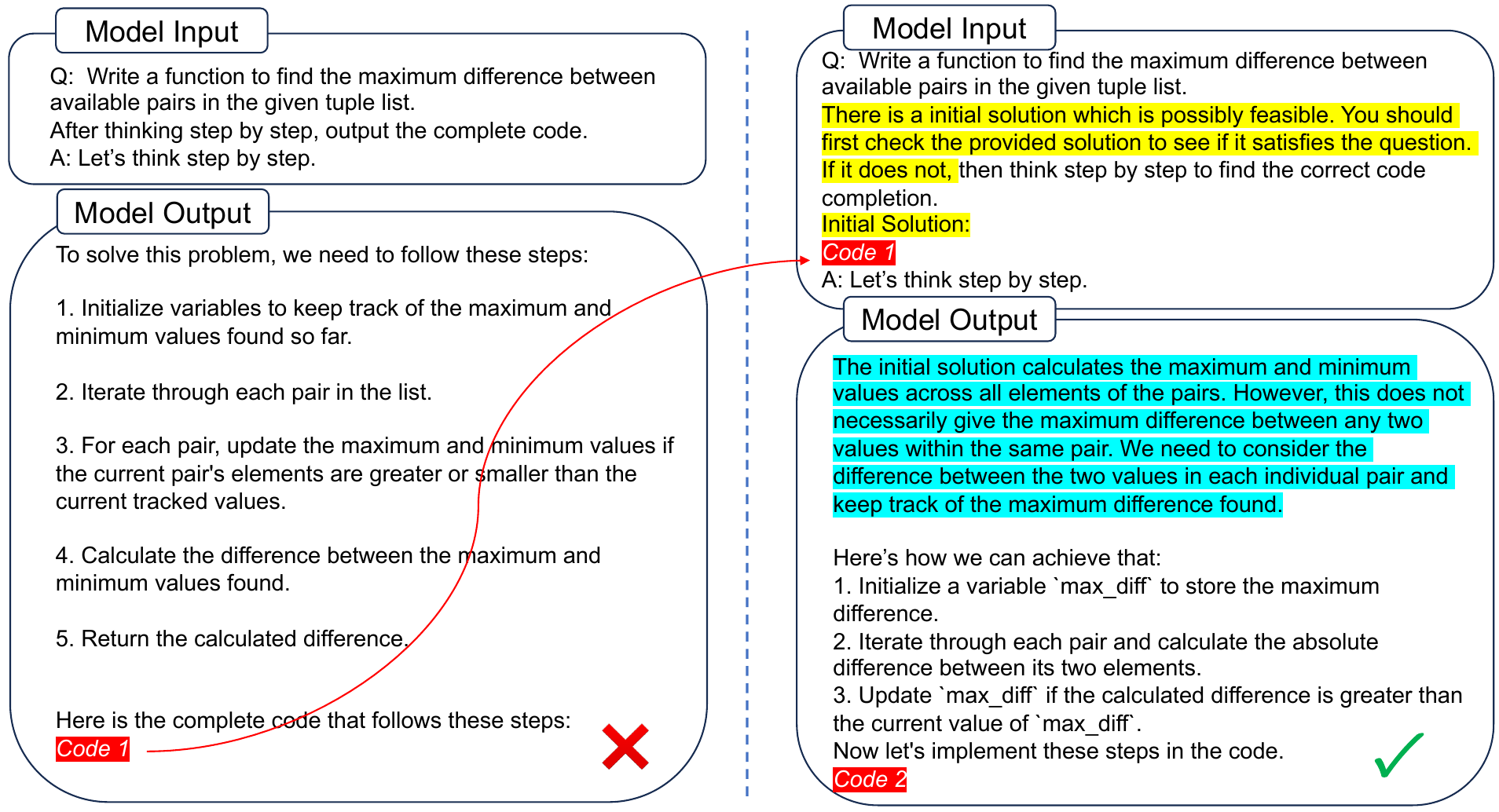}}
			\subfigure[Iter-VF for test-time scaling.\label{fig:iter}]{
				\includegraphics[width=0.7\textwidth]{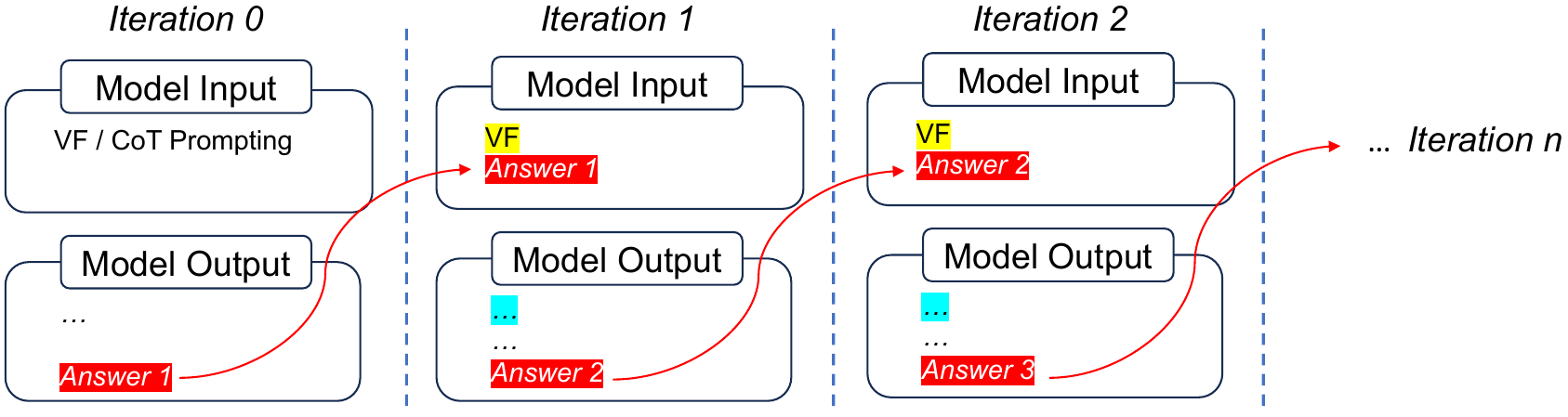}}
			\vspace{-10pt}
			\caption{Illustration of  (a) VF prompting with previously generated answer, and iterating such process as (b) Iter-VF for test-time scaling. }
			\label{fig:illus}	
			\vspace{-10pt}
		\end{figure*}

		\begin{figure*}[t]
			\centering
			\includegraphics[width=0.7\textwidth]{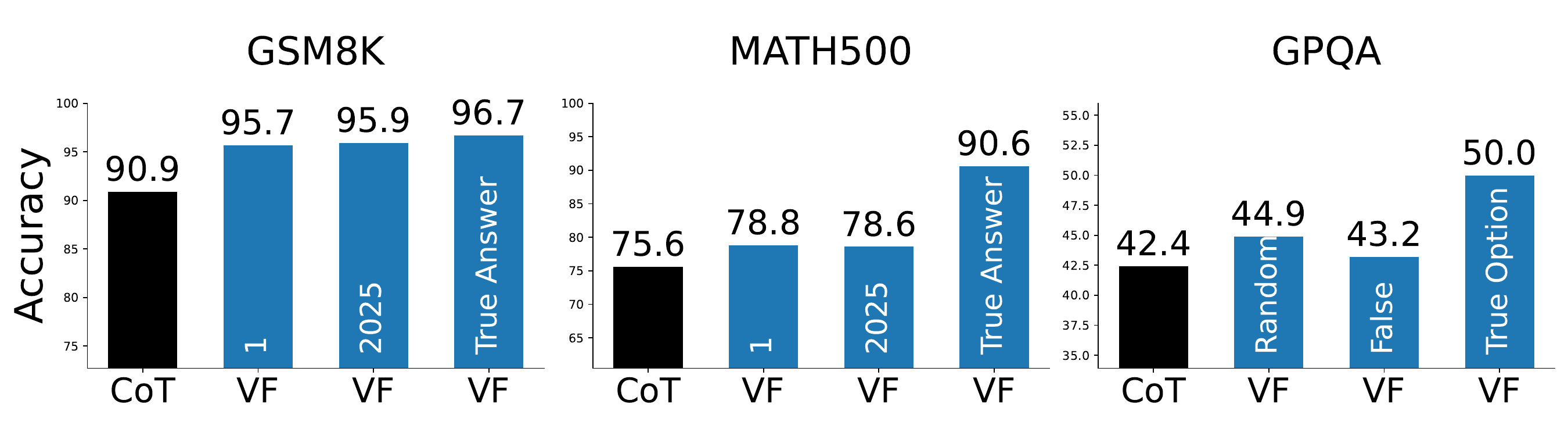}
			\vspace{-10px}
			\caption{Providing different answers to VF for verification. }
			\vspace{-10px}
			\label{fig:diff}	
		\end{figure*}

		\begin{figure*}[ht]
			\centering
			
			\includegraphics[width=0.99\textwidth]{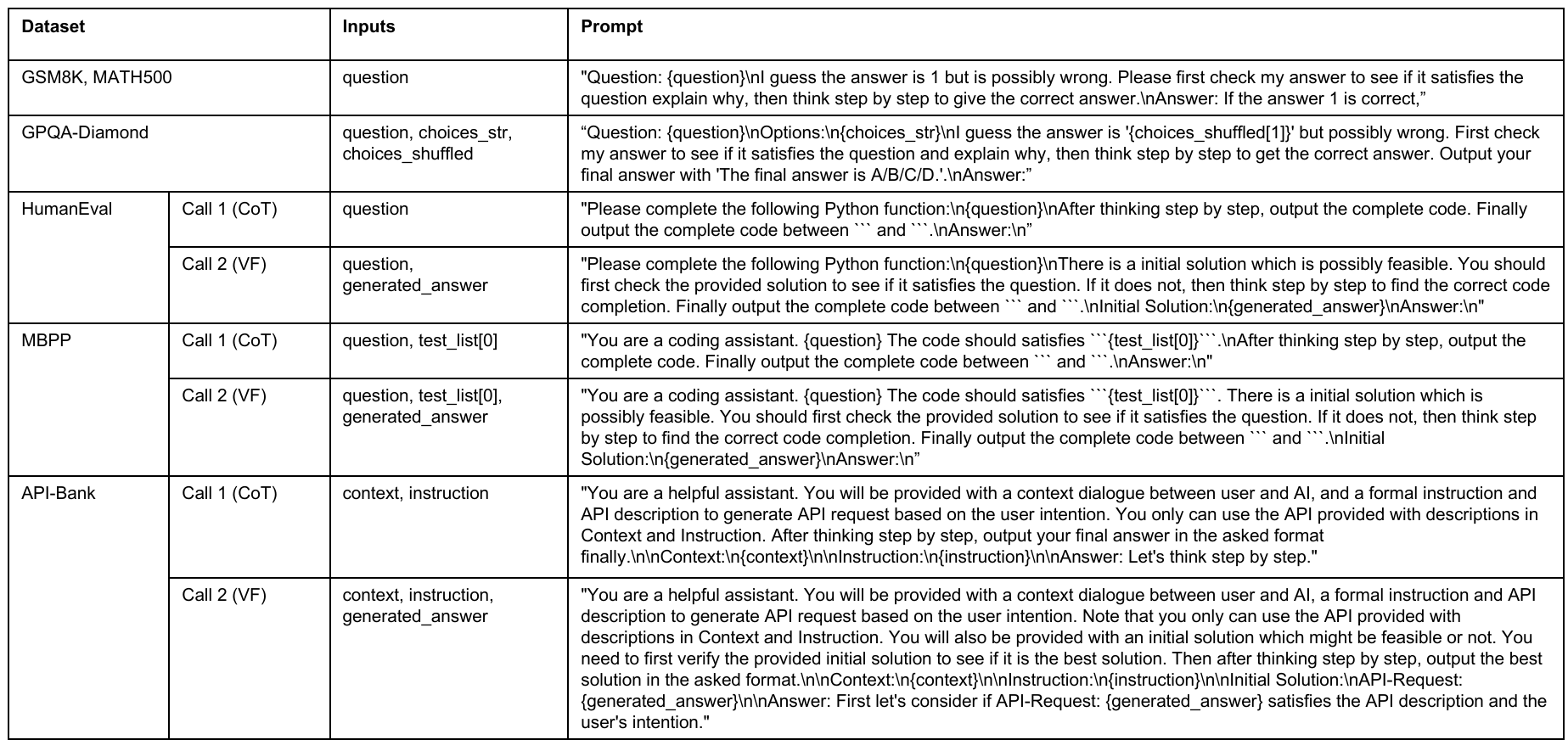}
			
			\caption{The specific prompts of VF for different datasets. }
			
			\label{fig:prompt}	
		\end{figure*}
		\begin{table*}[t]
			\centering
			\begin{tabular}{llcccccc}
				\toprule
				\multirow{2}{*}{\textbf{Model Family}} & \multirow{2}{*}{\textbf{(3.x-)Size (B)}} & \multicolumn{2}{c}{\textbf{GSM8K}} & \multicolumn{2}{c}{\textbf{MATH500}} & \multicolumn{2}{c}{\textbf{GPQA}} \\
				\cmidrule(lr){3-4} \cmidrule(lr){5-6} \cmidrule(lr){7-8}
				& & CoT & VF & CoT & VF & CoT & VF \\
				\midrule
				\multirow{4}{*}{Qwen2.5-Instruct}
				& 1.5 & 61.64 & \textbf{74.45} & 47.40 & \textbf{56.20} & -- & -- \\
				& 3   & 79.00 & \textbf{87.17} & 59.80 & \textbf{65.60} & 30.80 & \textbf{32.32} \\
				& 14  & 90.91 & \textbf{95.75} & 75.60 & \textbf{78.80} & 42.40 & \textbf{44.94} \\
				& 72  & -- & -- & -- & -- & 49.50 & \textbf{50.51} \\
				\midrule
				\multirow{4}{*}{Llama3-Instruct}
				& 3.2-1   & 49.58 & \textbf{60.12} & 23.00 & \textbf{31.00} & -- & -- \\
				& 3.2-3   & 74.13 & \textbf{80.18} & 42.20 & \textbf{48.60} & 31.41 & \textbf{32.83} \\
				& 3.1-8   & 84.15 & \textbf{88.60} & 45.60 & \textbf{52.60} & 27.10 & \textbf{30.30} \\
				& 3.3-70  & -- & -- & -- & -- & 43.43 & \textbf{47.98} \\
				\bottomrule
			\end{tabular}
			\caption{Comparison of accuracy (\%) between CoT and VF Prompting across Qwen2.5 and Llama3.x models on GSM8K, MATH500, and GPQA datasets.}
			\label{tab:vf_cot_comparison}
		\end{table*}

		\section{Distinguishing Iter-VF via Test-Time Scaling Update Dynamics }
		\label{sec:framework_differentiation}
		As a sequential TTS strategy, Iter-VF seems similar with existing strategies Self-Correction / Refine / Reflexion in the iterations. 
		In fact they are very different.
		Iter-VF distinguishes itself by (i) maintaining a Markovian process across iterations, that avoids context overflow and error accumulation; and (ii) instructing to do reasoning from scratch in every iteration. 
		
		Sequential TTS strategies can be expressed as an iterative procedure. Let $\bQ$ be the problem formulation and let $\bS_t$ denote the method's carried state at iteration $t$. A generic test-time method can be written as an update operator $\mathcal{T}$:
		\begin{equation}
			\bS_{t+1} = \mathcal{T}(\bQ, \bS_t) \label{eq:update_operator}.
		\end{equation}
		Different methods primarily differ in (i) what information is stored in the state $\mathbf{S}_t$, and (ii) how the operator $\mathcal{T}$ uses that state. This perspective allows us to categorize sequential TTS methods into two fundamental algorithmic families, summarized in Table~\ref{tab:tts_dynamics}: Trace-Editing Refinement (TER) and Answer-Hypothesis Testing with Re-Solve (AHTRS).
		
		\begin{table*}[t]
			\centering
			\small
			\vspace{-10px}
			\begin{tabular}{p{0.2\linewidth}| p{0.38\linewidth} |p{0.38\linewidth}}
				\toprule
				\textbf{Family} &Trace-Editing Refinement (Existing) & Answer-Hypothesis Testing with Re-Solve (Ours) \\
				\midrule
				\textbf{Representative Methods} & Self-Correction, Self-Refine, Reflexion & {Iter-VF} \\
				\addlinespace
				\textbf{State ($\mathbf{S}_t$)} & $\mathbf{S}_t = (\mathbf{A}_t, \mathbf{R}_t)$ (Answer + prior trace) & $\mathbf{S}_t = \mathbf{A}_t$ (Candidate answer only) \\
				\addlinespace
				\textbf{Update Operator ($\mathcal{T}$)} & Explicitly critiques and edits the earlier trace. & Verifies answer, then re-solves from scratch. \\
				\addlinespace
				\textbf{Context Window} &  $|\mathbf{S}_t| \sim \mathcal{O}(t)$. Grows continuously with the iteration budget. & $|\mathbf{S}_t| \sim \mathcal{O}(1)$. Remains compact and constant. \\
				\bottomrule
			\end{tabular}
			\vspace{-10px}
			\caption{Comparison of Sequential TTS Dynamics distinguishing Trace-Editing Refinement methods from our proposed Answer-Hypothesis Testing with Re-Solve approach.}
			\label{tab:tts_dynamics}
			\vspace{-5px}
		\end{table*}
		
		While TER explicitly encourages the model to reuse and patch its own prior reasoning $R_t$, structural errors and spurious rationales can be easily reinforced. Furthermore, the context window grows continuously with the iteration budget, leading to potential context overflow and degraded attention allocation. We attribute the observed empirical results (Section~\ref{sec:exp-tts}) to such reason \cite{huang2023large}. 
		As AHTR, by decoupling the Update step $\mathcal{T}$ from the historical trace $\bR_t$, Iter-VF maintains a lower, Markovian dependence. It inherits the search-space restriction benefits described in Section~\ref{sec:prompting}, without the anchor of the possibly flawed reasoning path.
		
		\section{Sensitivity on Provided Answer to Verify}\label{app:sensitive}
		It is important to understand if the provided answer $\bA'$ in VF prompt makes influence. We show the effect of providing different answers to VF prompting in Figure~\ref{fig:diff}. For GSM8K and MATH500, we evaluate the effect of providing trivial answers "1" and "2025", and the true answer respectively. For GPQA-Diamond problems  which are single choice problems, we evaluate the effect of providing random option, false option, and the true option respectively.

		Obviously, though it is not a valid method in application,
		providing the true answer in VF prompting (without telling the correctness, asking the LLM to verify first and then give correct answer) significantly improves performance. Such result indicates LLMs' capacity on verification and the fact that verifying an answer is easier than generating the correct answer.
		
		Oracle true answers can provide extra information, but it is not a practically applicable method, only studied for understanding the effect. Among minimal-prior-knowledge candidates the dominant factor is whether verification elicits useful constraints rather than the exact candidate value.
		Most importantly, it can be observed that providing different answers with minimal prior knowledge would not effect the final performance of VF prompting a lot.

		Note that for many problems, the answer space is not strict for LLMs to understand. As in MATH500 dataset, while many problems should not be answered with a scalar, the LLM can still interpret and verify the provided answer "1" with meaningful process.
		For example, the first test case of MATH500 is asking "Convert $(0,3)$ in rectangular coordinates to polar coordinates. $(r,\theta) ...$", and the LLM can response the VF prompting by "If the answer 1 is correct, ... the point \( (0, 3) \) in rectangular coordinates converts to polar coordinates \((r, \theta)\) with \( r = 1 \) ...". 
		Complex problems where it is hard to provide a random answer nontrivial to verify, will be studied in Section~\ref{sec:open}.


		\section{Prompts}\label{app:prompt}
		The table in Figure~\ref{fig:prompt} provides the specific prompts we use for evaluation in our experiments. One can also use the provided code including specific prompts and evaluation protocols.

			\section{Performance of CoT and VF with Random/Trivial Answer}\label{app:num}
			Table~\ref{tab:vf_cot_comparison} provides the results comparing CoT prompting and VF prompting. We evaluate LLMs from the Qwen2.5 \cite{qwen2} and Llama3 \cite{grattafiori2024llama} families, covering size from 1B to 72B: 
			Qwen2.5-1.5B / 3B / 14B / 72B-Instruct, Llama3.2-1B / 3B-Instruct, LLama3.1-8B-Instruct, Llama3.3-70B-Instruct.
			We use different x in Llama3.x, because different xs provide  models in different sizes and some sizes only exist in specific x.
				
			The performances of 1.5B/1B on GPQA-Diamond is not reported because they show performances almost indistinguishable with random guess (25\%). This means the benchmark is too hard for 1.5B/1B models that they show zero knowledge/reasoning ability, thus lack of comparison value. So we additionally evaluate the larger 72B/70B on GPQA-Diamond instead.

		\section{Error Case Analysis}
		\label{sec:error-analysis}
		Aggregate accuracy does not fully characterize how VF changes individual reasoning trajectories. We therefore inspect the overlap between CoT and VF outcomes by separating examples into four categories: both correct, VF-correct/CoT-wrong, VF-wrong/CoT-correct, and both wrong. The third category is especially important because it isolates cases where the verification-first instruction harms an otherwise correct CoT trajectory. Further inspection shows that such VF-wrong/CoT-correct cases do exist non-negligibly, especially for smaller models such as 3B or 8B models. Hence, VF should be understood as an average-case improvement mechanism rather than a per-instance monotonic guarantee over CoT.
		
		Based on the theoretical framework in Section~\ref{sec:prompting}, we categorize VF failures into three main modes. The first two correspond to failures of the two key steps required by the search-space restriction argument: detecting a constraint collision and then faithfully using the surfaced constraint in subsequent generation. The third mode concerns tasks where the main bottleneck is missing knowledge rather than missing constraint instantiation.
		
		\begin{table*}[t]
			\centering
			\small
			\begin{tabular}{p{0.25\linewidth}p{0.34\linewidth}p{0.33\linewidth}}
				\toprule
				\textbf{Failure mode} & \textbf{Mechanism} & \textbf{Representative symptom and implication} \\
				\midrule
				\textbf{Failure of Assumption~2: missed constraint collision} & The candidate answer violates one or more problem constraints, but the model fails to detect the decisive collision, or rejects the candidate only superficially without surfacing the violated constraint. & The verification trace does not introduce useful new constraints into the context. The subsequent generation therefore remains close to the original under-constrained CoT trajectory, so VF provides little or no benefit. \\
				\addlinespace
				\textbf{Failure of Assumption~3: incomplete verification-to-generation transfer} & The model detects that the candidate answer is wrong and may even verbalize relevant constraints, but it then misinstantiates, ignores, or incompletely reasons with these constraints during the final solving step. & In the house-flipping problem, VF rejects answer ``1'' but misinterprets ``increased the value by 150\%'' as the final value being $150\%$ of the purchase price, producing an incorrect profit instead of the correct \$70{,}000. In the omelet problem, VF rejects the trivial answer but double-counts the factor of three eggs, yielding 21 dozens instead of 7 dozens. These are genuine VF-induced errors, not merely failures to improve over CoT. \\
				\addlinespace
				\textbf{Knowledge-intensive rather than constraint-intensive tasks} & The limiting factor is missing domain knowledge rather than failure to attend to explicit problem constraints. In such cases, verifying a random or trivial candidate does not reliably recover the needed facts. & VF can still help when verification exposes a useful inconsistency, but the expected gain is weaker because search-space restriction cannot compensate for absent knowledge. This helps explain why gains on knowledge-intensive tasks such as GPQA-Diamond are smaller than on more constraint-rich mathematical problems. \\
				\bottomrule
			\end{tabular}
			\vspace{-5pt}
			\caption{Three representative VF failure modes aligned with the theoretical analysis. The first two correspond to failures of the verification-induced constraint restriction mechanism; the third occurs when the task is dominated by knowledge retrieval rather than explicit constraint satisfaction.}
			\label{tab:error_cases}
			\vspace{-10pt}
		\end{table*}
		
		These three modes share the same underlying source: limited model capability. VF introduces an additional verification-first requirement before the final answer is generated. When the model has sufficient verification, reasoning, and knowledge capacity, this extra requirement helps by surfacing constraints and restricting the logical search space. However, when the model is weaker, the same requirement can impose extra reasoning burden. Following the verification instruction may consume capacity or attention and may distract the model from a simpler forward reasoning path that CoT would have taken correctly. This explains why VF-wrong/CoT-correct cases can appear more frequently in smaller models, even though VF improves aggregate performance in our main experiments.
		
		Thus, VF failures are not a separate contradiction of the proposed mechanism, but boundary cases where one of the required model abilities is insufficient: detecting the constraint collision, completing the verify-then-solve reasoning process under the surfaced constraints, or possessing the background knowledge needed for verification. This also suggests practical safeguards. For reliability-critical applications, one can compare CoT and VF outputs and accept the answer only when they agree, use an external checker when available, or apply a lightweight consistency vote. For Iter-VF, early stopping after answer stabilization can reduce unnecessary re-solving and mitigate over-correction.
		
		\section{Test-Time Scaling Baselines}\label{app:tts}
		we compare Iter-VF with following TTS strategies: 
		\begin{itemize}[leftmargin=*]
			\item Self-Correction \cite{madaan2023self,shinn2023reflexion,muennighoff2025s1}: A sequential strategy that ask LLM to reflect and refine output iteratively. 
			
			\item PHP \cite{zheng2023progressive}: A sequential strategy that concatenates all previous answers after the input question by"(Hint: The answer is near
			to $a_1,\cdots,a_{t-1}$)". 
			
			\item  Self-Consistency \cite{wang2022self}: A parallel strategy that generates multiple reasoning paths leading to answers independently, and determines the final answer by majority-voting.
			
			\item Best-of-N \cite{lightman2023let,yao2023tree}: A parallel strategy that multiple reasoning paths leading to answers independently, and asks the LLM to evaluate the quality of each one respectively, and determines the final answer with highest score.
		\end{itemize}
		Note that while some methods \cite{lightman2023let,yao2023tree,shinn2023reflexion,muennighoff2025s1} could require training an additional model as evaluator, and use it to progressively evaluate decomposed reasoning steps, or obtaining external feedback, they are reduced to such implementation for fair comparison under the no-training and no task-specific prior knowledge setting.

	\end{document}